\RequirePackage{amsmath}
\documentclass[smallcondensed]{svjour3}
\usepackage[T1]{fontenc}
\usepackage[latin9]{inputenc}
\usepackage{float}
\usepackage{fancybox}
\usepackage{calc}
\usepackage{amsmath}
\usepackage{amssymb}
\usepackage{hyperref}
\usepackage[title]{appendix}
\usepackage[justification=centering]{caption}
\usepackage{enumitem}
\usepackage{caption}
\usepackage{subcaption}

\setlist{topsep=0pt,after=\newline} 

\usepackage{graphicx}
\makeatletter

\floatstyle{ruled}
\newfloat{algorithm}{tbp}{loa}
\providecommand{\algorithmname}{Algorithm}
\floatname{algorithm}{\protect\algorithmname}

\makeatother

\usepackage[english]{babel}
\begin{document}
\title{Level-set KSVD}

\author{Sapir, Omer  \and Axelrod, Michal \and Niv, Ariela  \and  Klapp, Iftach \and  Sochen, Nir}
\institute{Sapir, Omer \and Sochen, Nir
\at Department of Applied Mathematics, Tel Aviv University, Tel Aviv 69978, Israel
\and Axelrod, Michal \and Niv, Ariela
\at The Israel Cotton Board Ltd.
\and Klapp, Iftach \and Sapir, Omer
\at Information and Mechanization Engineering, Institute of Agricultural Engineering, Agricultural Research Organization, Volcani Institute, Bet Dagan, Israel \\\email{iftach@volcani.agri.gov.il}
}

\maketitle
\begin{abstract}
We present a new algorithm for image segmentation - \textbf{Level-set KSVD}. Level-set KSVD merges the methods of sparse dictionary learning for feature extraction and variational level-set method for image segmentation. Specifically, we use a generalization of the Chan-Vese functional with features learned by KSVD.
The motivation for this model is agriculture based. Aerial images are taken in order to detect the spread of fungi in various crops. Our model is tested on such images of cotton fields. The results are compared to other methods.

\end{abstract}

\section{Introduction}

Image segmentation is a basic task in image analysis and computer vision. It is done under the assumption that two regions are distinguishable from one another by the fact that pixels (or neighborhoods thereof) in different regions of the image are samples from different Probability Distribution Functions (PDFs). The samples may be the pixels' values directly or local features attached to the pixels.  Otherwise no human and no algorithm can distinguish between the regions. Early attempts for the formulation of segmentation algorithms were devised under the assumption that the main feature to consider is the pixels' value. The algorithm of choice in this regard is the Chan-Vese algorithm \cite{Chan2001} that assumes two different regions (e.g. foreground and background) such that the grey values in the two regions are realizations of two normal distributions with different expected values and the same standard deviation.  These naive assumptions were later changed for a more realistic PDFs in the two regions by Cremers et al \cite{cremers2007review}. 

In the next wave of image processing techniques the idea of sparsity was applied as a prior together with feature/dictionary learning based on a database of relevant images for the task. Sparsity was used more for the restoration of images (denoising, debluring etc.) than for segmentation. The main issue here, besides the sparsity, is learning the relevant features of {\em{a neighborhood}} of a pixel. These features form a dictionary such that each neighborhood can be well approximated as a linear combination of just few of the atoms in the dictionary. The main conceptual importance for us is that features/atoms are learned and not devised by hand.   

Modern approach using deep learning gives up modelling the problem and focus on finding a huge parametric function that approximates directly the function that maps the input image to the labeled image. Unlike earlier approached this necessitates a large annotated (manually segmented) database of relevant class of images \cite{garcia2017review}. 

In this paper we try to stay with a small non-annotated database and combine the naive Chan-Vese approach with the feature learning technique of sparsity \cite{Aharon2006}.  Our suggested Level-set KSVD algorithm merges the methods of sparse dictionary learning for feature extraction and the variational level-set method for image segmentation. Specifically, we use (a generalized form of) the Chan-Vese functional with features extracted by KSVD.

The main assumption is that the patches in each region are well approximated by a dictionary that is devoted to that region. The signals in the different segments of the image are assumed to be compressible in the sense that the residue of their description by the dictionary is a random variable i.e. noise that is a realization, at each patch of the $i$-th region, of a normal distribution with expectation $\mu_i$ and covariance $\Sigma_i$. The dictionary is learned from only {\em one image}! The expectation and covariance are adapted along the flow.  The model was tested on real-life agricultural aerial images to detect spread of fungi in cotton fields and was compared to other methods.



Our main novelty is in the combination of the following: 

\begin{itemize}
    \item Using dictionary learning for automated feature/texture extraction.
    \item Combining the `neighborhood' features (patch window) extracted by KSVD with the local (curvature, gradient) and global (mean, variance) features in the Chan-Vese functional.
    \item Extending Chan-Vese to multi-variate distributions with covariances.
    \item Helping to solve significant agricultural issues with real-life data.
\end{itemize}

In the next subsection we give the context in which this work was realized. This includes the real-life problem, review of the relevant techniques used in our algorithm and the main works that were previously done in this application domain and with which we compared our algorithm.

\section{Motivation, Review and previous work}

This subsection includes several parts: First, a short description of the motivation to our work for the detection of the \textit{Macrophomina} fungus in the agriculture setting with its commercial and agricultural importance is given. Second, a short review of previous works on low-altitude agricultural imagery anomaly detection, including some of our early attempts is presented. Finally, for the sake of being self-contained, we provide a short reminder of the basic building blocks of our algorithm, namely the \textbf{Chan-Vese} functional and the \textbf{KSVD} algorithm. 

\subsection{\label{sec:Anomalies-in-agricultural}Anomalies in agricultural crops}

Agricultural fields and crops are disrupted by many phenomena, including insects and diseases. This work focuses on detecting anomalies caused by \textit{macrophomina phaseolina} 

\label{macrophomina}
\textit{Macrophomina phaseolina} is a fungus that infects hundreds of species \cite{farr1989fungi}, most notably - important agricultural crops. It is a harmful pathogen with symptoms such as wilting, premature dying and loss of vigor. Recent years have seen an outburst of \textit{Macrophomina} in cotton around the world \cite{omar2007diversity} \cite{baird1999first} \cite{kaur2012emerging}, heavily damaging fields and reducing yield. It is important to detect the fungus spread as soon as possible and to be able to estimate the spread and damage caused by it. Visually, the fungus creates patches of wilted and dead plants, seen from the air as brown spots in the green field. Ideally, aerial imagery supplemented with segmentation algorithm can be used for early detection and spread estimation.

\subsection{\label{sec:Aerial-imagery-in-agriculture}Aerial imagery in agriculture}

Major technological developments in recent years have made Unmanned Aerial Vehicles (UAVs) into a reliable and relatively cheap commercial commodity for both the scientific community and individual farmers \cite{Ajayi20} \cite{veroustraete2015rise}. Innovations in photography allowed for the commercialization of high-quality low-cost and low-weight digital cameras. Combined together, UAV mounted cameras provide an accessible solution for taking low-altitude high-resolution aerial imagery of crop fields \cite{kulbacki2018survey} \cite{malveaux2014using}. \par
This advancement can provide farmers better monitoring on a regular basis by taking aerial images whenever necessary \cite{shahbazi2014recent}. Thus small anomalies (such as early outburst of diseases) are now visible, motivating the development of new segmentation algorithms to detect them.

\subsection{Image processing for agriculture}

Image processing for agriculture tasks is a well studied subject as reviewed in \cite{crommelinck2016review}. Specifically, the task of image segmentation is surveyed in \cite{jayanthi2017survey}. Remote sensing segmentation uses satellite or high-altitude imagery for vegetation estimation  \cite{geipel2014combined}  or large-scale stress detection. These works provide good overview regarding entire fields \cite{crommelinck2016review}. Another application for agricultural image segmentation is monitoring rangeland \cite{laliberte2010acquisition}.
Some works focus on separating plants and soil \cite{hamuda2016survey} \cite{poblete2017detection} \cite{hernandez2016optimal}, and weed detection (mostly by finding the crop rows or extracting features describing crops and weeds \cite{lottes2017uav}, \cite{laliberte2010acquisition}).

During our research we tried different approaches including: thresholding \cite{Otsu1979}, Color space transformation (successful in specific cases), regular Chan-Vese  \cite{Chan1999}, vector-valued Chan-Vese \cite{Vese2002} and texture-based Chan-Vese \cite{sandberg2002level} , Constrained Gaussian mixture model (CGMM) \cite{Greenspan2006}, Variational CGMM \cite{Freifeld2007}, as well as a new approach recently published by us \cite{klapp2017localization} that uses the effects of blurring on the image histogram. 

\subsection{\label{sec:Chan-Vese-Algorithm}Chan-Vese Algorithm}

Chan and Vese suggested a segmentation algorithm \cite{Chan1999} that partitions an image by assuming that the intensity levels in each meaningful segment follow a normal distribution. The algorithm assumes smoothness of the boundary between the foreground and background. Therefore, it is possible to classify each point in the image to a segment it belongs with Maximum A-posteriori Probability (MAP) method. Essentially, it transforms the segmentation task of an entire image to a classification task of each and every point. To encourage desirable properties, such as connectedness of objects and smooth curves, the algorithm rewards and penalizes for these patterns.
The basic Chan-Vese assumes a normal distribution of the grey values of the pixels in a region with the average intensity as mean and with a-priori unit variance. It is readily generalized as multi-variate distribution for RGB images \cite{Vese2002}. Other generalizations expand and extend this probabilistic notion. Section \ref{sec:Probabilistic-view} offer an analysis of our proposed method from this perspective. \par
The mathematical form of basic Chan-Vese is a functional with 4 terms:
\begin{enumerate}
    \item The curvature of the contour
    \item The probability of belonging to the foreground
    \item The probability of belonging to the background
    \item The total foreground area 
\end{enumerate}

For defining the functional we will define some terms.
Let $H\left(z\right)$ be the Heaviside function defined as:
\begin{equation}
H\left(z\right)=\begin{cases}
1 & z>0\\
0 & z<0
\end{cases}
\label{eq:Heaviside}
\end{equation}
With derivative $\delta\left(z\right)=\frac{d}{dz}H\left(z\right)$

Let $\phi\left(x,y\right)$ be a level-set function (As defined in \cite{Osher1988}) for segmentation, such that
\begin{equation}
\phi\left(x,y\right)=\begin{cases}
\phi\left(x,y\right)>0 & \left(x,y\right)\in foreground\\
\phi\left(x,y\right)<0 & \left(x,y\right)\in background 
\end{cases}
\end{equation}

The functional is defined as follows:

\begin{align}
F_{\mu,\nu,\lambda_{1},\lambda_{2}}\left(\phi,c_{1},c_{2}\right)  = &\underset{\Omega}{\int}\Big[\mu\delta\left(\phi\left(x,y\right)\right)\cdot\left|\nabla\left(\phi\left(x,y\right)\right)\right|\nonumber \\ 
& +\lambda_{1}\left|\left(I\left(x,y\right)-c_{1}\right)\right|^{2}\cdot H\left(\phi\left(x,y\right)\right)\nonumber \\  
& +\lambda_{2}\left|\left(I\left(x,y\right)-c_{2}\right)\right|^{2}\cdot\left(1-H\left(\phi\left(x,y\right)\right)\right)\label{eq:chan-vese-edited}\\  
& +\nu H\left(\phi\left(x,y\right)\right)\Big]dxdy\nonumber 
\end{align}

where $\mu,\nu,\lambda_{1},\lambda_{2}$ are weight parameters configured empirically; $c_{1}$ and $c_{2}$ are average intensities of foreground and background, respectively; $\Omega$ is the image region on the image plane; $I$ is the image intensity; and  $\left(x,y\right)$ are coordinates in the image.

Each term gives a weighted penalty. The first term penalizes long and complex contours. The second and third terms control variances of the foreground and background regions, respectively. For instance,  $\frac{\lambda_{1}}{\lambda_{2}} > 1$ forces a foreground with low-variance in intensity. The fourth term penalizes large objects (or encourages them if $\nu<0$).

\subsection{\label{sec:KSVD-Algorithm}KSVD Algorithm}

The KSVD algorithm \cite{Aharon2006,Elad2006} is a method for learning a dictionary to sparsely represent a given set of $N$ signals $\{x_i\}_{i=1}^{N}$ (sometime called samples). Dictionary learning consists of finding an over-complete basis for the signal space such that each is approximated by a linear combination of only a few elements from the dictionary. Formally, assuming all signals are vectors in $\mathbb{R}^{d}$ for some integer $d$, the dictionary learning goal is to solve the following:

\begin{equation}
\min\limits _{D,A}\{\|M-DA\|_{F}^{2}\}\quad\text{s.t. }\quad\forall i\;,\|\alpha_{i}\|_{0}\leq\rho
\label{eq:ksvd}\end{equation}
where $M$ is the signal matrix in which the i-th column is the signal $x_i$; $D$ is the learned dictionary, whose columns, a.k.a - atoms, form an over-complete basis; $\alpha_i$ is the coefficient vector of the sparse representation of the signal $x_i$;   $A$ is the sparse representation coefficient matrix (or simply the coefficient matrix, as we name it below), whose columns are $\{a_i\}_{i=1}^{N}$; $\rho$ is the sparsity factor controlling the maximum amount of atoms permitted in each representation; $\|\alpha_{i}\|_{0}$ is the $0$-th norm which denotes the number of non-zero elements in $\alpha_{i}$; $\|M-DA\|^{2}$ is the approximation error, $\|B\|_{F}^{2}$ denotes the Frobenius norm defined as $\sqrt{\underset{i,j}{\sum}B_{ij}^{2}}$.

Since its release, many variations and improvements have been introduced such as color KSVD \cite{Mairal2008}, multi-scale KSVD \cite{Mairal2008b} and LC-KSVD \cite{Jiang2011}, Discriminative KSVD \cite{Zhang2010}, Analysis KSVD \cite{Rubinstein2013}, as well as an efficient Matlab implementation \cite{Rubinstein2008}. \par

\subsubsection{\label{sec:Patches}Patches}

\begin{figure}[H]
\noindent \centering
\includegraphics[scale=1]{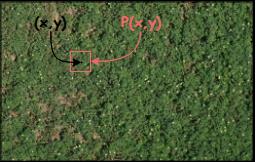}
\par
\caption{\label{fig:image-patch-illustration}Illustration of pixel and patch notation in an image.}
\end{figure}

 In Level-set KSVD, each pixel is assigned a patch (cropped from the image) as a feature vector. As shown in Figure \ref{fig:image-patch-illustration}), the pixel, marked by a black arrow, located at position $(x,y)$, induces the red patch $P(x,y)$ which describes a small area around the pixel. The atoms are learned from those patches. During the learning phase, the atoms capture essential characteristics for the immediate surrounding of different pixels. 

This enables assigning `neighborhood' features to a single pixel in addition to local features (such as intensity level and gradient) and global region-based features (such as mean and variance). \par

\subsubsection{\label{sec:Pursuit Algorithm}Pursuit Algorithm}

The sparse representation coefficient matrix $A$ is calculated by a pursuit algorithm such as Matching Pursuit (MP) \cite{mallat1993matching}, Orthogonal Matching Pursuit (OMP) \cite{pati1993orthogonal}, or Basis Pursuit (BP) \cite{chen1994basis}. \\
Finding the optimal sparse representation for a given signal requires checking all atom combinations for a given dictionary. which is NP-hard and not feasible. Therefore, Matching pursuit (MP) was introduced in \cite{mallat1993matching} to find a computationally-reasonable sub-optimal solution. 
Matching pursuit (MP) is a greedy algorithm to find the approximate projection of a signal onto the span of an over-complete dictionary. 

\subsubsection{\label{sec:ROC-Curve}ROC Curve}

A Receiver Operating Characteristic curve (ROC curve), is a graphical plot to show the strength of a binary classifier at different threshold values. The Y axis features the true positive rate (TPR), while the X axis presents the false positive rate (FPR). 
Since both axes represent rates, they have a range from 0 to 1. 0 is the best possible FPR, 1 is the best possible TPR, therefore the point (0,1) represent the perfect classifier.
The linear line $x=y$ from (0,0) to (1,1) represent a random classifier.
A point on the graph is calculated by setting a classification threshold and checking TPR and FPR for this threshold. The graph is built by taking many threshold values. 
The area under the curve (AUC) range from 0 to 1. The Ideal classifier has an AUC of 1. random classifier has an AUC of 0.5.

\subsection{Related previous work}


The most similar work to our new algorithm is by Mukherjee and Acton \cite{mukherjee2014region} who suggested 
using Legendre polynomials in the Chan-Vese functional in order to capture intensity variations in both foreground and background segments. While their work is close to the proposed method, it suffers from three drawbacks. First, it assumes prior knowledge about the objects, which is, in general, not always known or available and not applicable to our dataset. Second, it uses static pre-determined basis functions. Third, it tries to approximate the entire image, which potentially lower its local accuracy. Our proposed method tackle the problem in a new fashion, the estimation done locally using local patches for more detailed feature extraction.


\section{\label{sec:model}The proposed algorithm: level-set KSVD}

The proposed model is composed of learning two dictionaries $D_{1}$ and $D_{2}$ for the foreground and background, respectively. This is followed by segmenting the image, pixel-by-pixel, with a special functional that uses the dictionaries as extracted features. It is important to note that only a single image is required for creating the dictionaries. Afterwards, images with similar features can be segmented without further learning. \\
Formally, a patch $P\left(x,y\right)$ of an image $\Omega$ is a small fixed-size image cropped from a large image, centered at point $(x,y)$ (as illustrated in Figure \ref{fig:image-patch-illustration}). \\
Let $\Omega_{D}\left(x,y\right)$ be an image which is a function on the $\left(x,y\right)$-plane that is used for dictionary learning. Assume $\Omega_{D}\left(x,y\right)$ is composed of two segments - anomalies (foreground) and non-anomalies (background). We can classify each patch in the image to either foreground or background. When applied to all patches in the image, this patch-classification is equivalent to a pixel-by-pixel segmentation of the image. \\
Let $M_{1}=\left\{ P_{n}\left(x,y\right)\right\} _{n=1}^{N_{1}}$ be a set of $N_{1}$ patches of similar size, belonging to the foreground and let $M_{2}=\left\{ P_{n}\left(x,y\right)\right\} _{n=1}^{N_{2}}$ be a set of $N_{2}$ patches of similar size, belonging to the background. These sets of patches are annotated manually for a single image. \\
Let $D_{1}=\left\{ D_{1k}\right\} _{k=1}^{K_{1}},D_{2}=\left\{ D_{1k}\right\} _{k=1}^{K_{2}}$ be two dictionaries, of sizes $K_{1},K_{2}$ atoms, respectively, each trained by $M_{1},M_{2}$. \\
Let $\alpha_{1}\left(x,y\right)=\left\{ \alpha_{1k}\right\}_{k=1}^{K_{1}},\alpha_{2}\left(x,y\right)=\left\{\alpha_{2k}\right\} _{k=1}^{K_{2}}\subset\mathbb{R}$ be the sparse representation of some patch $P\in M$ with $D_{1},D_{2}$ respectively. \\
Let $A_{1},A_{2}$ be sparse representation matrices of the N patches, such that$\left\Vert M_{1}-D_{1}A_{1}\right\Vert $ and $\left\Vert M_{2}-D_{2}A_{2}\right\Vert $ are minimal w.r.t. some sparsity constraint ( In other words - $M_{1},M_{2}$ are the training sets of $D_{1},D_{2}$ respectively). \\
We now proceed to the design of the segmentation functional.

Following the above, by using the dictionaries as prior in the Chan-Vese model, an image $\Omega_{S}\left(x,y\right)$ can be segmented by minimizing the following functional:
\begin{align}
F_{\mu,\nu,\lambda_{1},\lambda_{2}}\left(\phi,\alpha_{1},\alpha_{2}\right) & =\underset{\Omega_{S}}{\int}\Big[\mu\delta\left(\phi\left(x,y\right)\right)\cdot\left|\nabla\left(\phi\left(x,y\right)\right)\right|+\nu H\left(\phi\left(x,y\right)\right)\nonumber\\
 & +\lambda_{1}\left\|P\left(x,y\right)-D_{1}\alpha_{1}\left(x,y\right)\right\|^{2}H\left(\phi\left(x,y\right)\right)\label{eq:first-functional}\\
 & +\lambda_{2}\left\|P\left(x,y\right)-D_{2}\alpha_{2}\left(x,y\right)\right\|^{2}\left(1-H\left(\phi\left(x,y\right)\right)\right)\Big]dxdy\nonumber
\end{align}

See section \ref{sec:Chan-Vese-Algorithm} for definitions of $H\left(z\right)$ , $\delta\left(z\right)$ and $\phi\left(x,y\right)$ \par

The model can be understood intuitively. Each point $(x,y)\in\Omega$ is assigned as either ``inside $C$'' or ``outside $C$'' , where $C$ is the contour line defined by $\phi\left(x,y\right)=0$, so that the indicator function of ``inside $C$'' is  $H(\phi)$ and the indicator function of ``outside $C$'' is  $1-H(\phi)$. The assignment depends on the approximation by $D_{1}$ and $D_{2}$, as well as on the constraints on the contour's curvature and the area it encloses. \par
The functional (Eq. \eqref{eq:first-functional}) can be minimized for $\phi$ using the Euler-Lagrange equation with a numerical solver (see Section \ref{sec:Euler-Lagrange-equation} below). \par
The sparse representation coefficient vectors $\alpha_{1},\alpha_{2}$ are calculated by a chosen pursuit algorithm.
Since the coefficient vectors depend only on $P$ and $D_1, D_2$, they can be calculated before running numerical minimization. This replaces the classical dual-step numerical approach for Chan-Vese with a simpler and faster scheme.

\subsection{\label{sec:Probabilistic-view}Probabilistic view}

The probability density function of a normal distribution of a random variable $X$ with mean $\mu$ and standard deviation $\sigma$ is:
\begin{equation}
{\frac {1}{\sigma\sqrt {2\pi }}}\;e^{-\frac{1}{2}\left(\frac{X-\mu}{\sigma}\right)^{2}}
\label{eq:normal-pdf}\end{equation}

The basic Chan-Vese functional implicitly assumes that the intensity of pixels in the foreground is distributed normally with a mean of $c_{1}$ and a standard deviation of $\sigma_{1}$. $\sigma_{1}$ relates to the training parameter $\lambda_{1}$ as follows (See Appendix \ref{appendix:chan-vese-log-likelihood} for details):
\begin{equation}
\lambda_{1}=\frac{1}{\sqrt{2\pi\sigma_{1}}}
\label{eq:lambda-sigma}
\end{equation}

We will now use this connection to re-define the fidelity terms of our functional. Let the object approximation error of the foreground be:
\begin{align}
e_{1}\left(x,y\right)=P\left(x,y\right)-D_{1}\alpha_{1}\left(x,y\right) \label{eq:approx-error}
\end{align} \par
We extend our model to assume $e_{1}$ is a multi-variate random vector that is normally distributed as per the regular probability density function with some mean $\mu_{1}$ and (unknown) covariance matrix $\Sigma_{1}$:
\begin{equation}
{\displaystyle {\begin{aligned}{\frac {e^{-{\frac {1}{2}}({\mathbf {e_{1}} }-{\boldsymbol {\mu_{1} }})^{\mathrm {T} }{\boldsymbol {\Sigma }_{1}}^{-1}({\mathbf {e_{1}} }-{\boldsymbol {\mu_{1} }})}}{\sqrt {(2\pi )^{k}|{\boldsymbol {\Sigma_{1} }}|}}}\end{aligned}}}
\end{equation}
where $k$ is the dimensionality of $e_{1}$ which is equal to the number of pixels per patch $P\left(x,y\right)$. \par
Our dictionaries provide good representation for each patch belonging to their class. We found, empirically, the error to have zero mean ($\mu_{1}=0$). Using this result, we get the following foreground term in our functional:
\begin{equation}
\label{eq:stat-approach}
\left\|P(x,y)-D_{1}\alpha_{1}(x,y)\right\|^{2}_{\Sigma_{1}}=\left(e_{1}-\mu_{1}\right)^{T}\Sigma_{1}^{-1}\left(e_{1}-\mu_{1}\right)=e_{1}^{T}\Sigma_{1}^{-1}e_{1}
\end{equation}
Note: The parameter $\lambda_1$ was dropped. This is because it represents the variance of $e_{1}$ (see Eq. \eqref{eq:lambda-sigma}) and therefore it is already included in the covariance matrix $\Sigma_{1}$. \par
We assume all terms in the approximation error have equal importance and weight. Therefore we force them to have variance of $1$, and scale the covariance terms in $\Sigma_{1}$ accordingly. This results in using the Correlation matrix of the approximation error (see Appendix \ref{appendix:correlation-matrix} for details). We get following term:

\begin{equation} 
\left\|P(x,y)-D_{1}\alpha_{1}(x,y)\right\|^{2}_{C_{1}}=e_{i}^{T}C_{1}^{-1}e_{1}
\end{equation}
where $C_{1}$ is the Correlation matrix of $e_{1}$. \par

Following the same process, the background fidelity term is:
\begin{equation} 
\left\|P(x,y)-D_{2}\alpha_{2}(x,y)\right\|^{2}_{C_{2}}=e_{2}^{T}C_{2}^{-1}e_{2}
 \end{equation}
 
As mentioned earlier, the approximation error $e_{1}(x,y)$ belongs to a single distribution for all pixels. Therefore, we use the same correlation matrix $C_{1}$ for all foreground pixels (and respectively  $C_{2}$ for all background pixels).

Following this probabilistic view, we implement these statistical terms in our functional as follows:

\noindent\doublebox{\begin{minipage}[t]{1\columnwidth - 2\fboxsep - 7.5\fboxrule - 1pt}%
\begin{align}
\noindent
F_{\mu,\nu}\left(\phi,e_{1},e_{2}\right) & =\underset{\Omega}{\int}\mu\delta\left(\phi\right)\cdot\left|\nabla\left(\phi\right)\right|+\nu H\left(\phi\right)\label{eq:main-cvd}\\
 & +e_{1}^{T}C_{1}^{-1}e_{1}H\left(\phi\right)+e_{2}^{T}C_{2}^{-1}e_{2}\left(1-H\left(\phi\right)\right)dxdy\nonumber 
\end{align}
\end{minipage}}
\newline \par
Removing the weight parameters $\lambda_{1},\lambda_{2}$, gives the 3rd and 4th terms $e_{1}^{T}C_{1}^{-1}e_{1}$ and $e_{2}^{T}C_{2}^{-1}e_{2}$ an effective constant weight of 1. This was tested and confirmed empirically (see Section \ref{sec:level-set-ksvd-framework} and Figure \ref{fig:Sample-segmentation-results}).
\par
Since the weight parameters are directly related the variance of object intensity, configuring them is equivalent to searching for this variance. In practice, this variance is unknown. Therefore, we incorporate the variance, through the Correlation Matrix, into our method. This is more robust and generalized since it allows us to find the variance during the minimization phase. It is also significantly faster compared to empirically searching for the correct parameters.

\subsection{\label{sec:Euler-Lagrange-equation}Euler-Lagrange equation}

Following basic Chan-Vese method, we minimize $F_{\mu,\nu}\left(\phi,e_{1},e_{2}\right)$ w.r.t $\phi$ to deduce the associated Euler-Lagrange equation for $\phi$. The Euler-Lagrange equation depends on the scalar product that is defined on the class of functions in which we are looking for a minimizer. In particular the relevant scalar product for the space of functions on the curve is  
\begin{equation}
    (f,g) = \int f(s)g(s)ds = \int f(x,y)g(x,y)\underbrace{|\nabla H(\phi(x,y))|dxdy}_{ds}
\end{equation}
Defining the descent direction by an artificial time $t\geq0$, the equation for the gradient descent solution $\phi\left(t,x,y\right)$ is \cite{Chan2001}:

\begin{align}
\frac{\partial\phi}{\partial t} & =\frac{1}{|\nabla\phi|}\left[\mu\text{div}\left(\frac{\nabla\phi}{\left|\nabla\phi\right|}\right)-\nu-\left(e_{1}^{T}C_{1}^{-1}e_{1}\right)\right.\nonumber \\
 & \left.+\left(e_{2}^{T}C_{2}^{-1}e_{2}\right)\vphantom{\left(\frac{\nabla\phi}{\left|\nabla\phi\right|}\right)}\right]\,in\,\left(0,\infty\right)\times\Omega,\label{eq:EL-grad-descent}\\
\phi\left(0,x,y\right) & =\phi_{0}\left(x,y\right)\,\,in\,\,\Omega\nonumber \\
\frac{\delta\left(\phi\right)}{\left|\nabla\phi\right|}\frac{\partial\phi}{\partial\overrightarrow{n}} & =0\,\,on\,\,\partial\Omega\nonumber 
\end{align}
See \cite{Sochen2022} for details of this derivation.

\section{Implementation details}

\subsection{Numerical Scheme }

Equation \eqref{eq:EL-grad-descent} can be readily solved numerically. We used central difference first-order and second-order derivatives of the level-set function $\phi(x,y,t)$ on a 3x3 stencil to calculate the curvature $\kappa$ using the formula (See \cite{Osher1988} for details): 

\begin{equation}
\kappa = \text{div}\left(\frac{\nabla\phi}{\left|\nabla\phi\right|}\right) = \frac{\phi_{xx}\phi_{y}^{2}-2\phi_{xy}\phi_{x}\phi_{y}+\phi_{yy}\phi_{x}^{2}}{(\phi_{x}^2+\phi_{y})^\frac{3}{2}}
\end{equation}
Our numerical method follows \cite{lankton2016}. It involves initializing $\phi$ with $\phi_{0}$, and evolving a narrow band around the curve. We also reinitialize $\phi$ every few iterations so we fixed the $\nabla\phi$ outside of the expression on the right to be 1. This simplifies the calculation and only slightly affects the velocity of the different level-sets in the narrow band and does not change the minimizer. In addition, we added a convergence test that checks the difference in area between consecutive steps. \par
For $\phi_{0}$, we chose a signed distance function (SDF) with the checkerboard mask, with small circles.
To insure numerical stability, we evolve only a proximal neighborhood around the curve, where $\left|\phi(x,y)\right|<\tau$. The parameter $\tau$ controls the extent of the curve evolution. We found the best results are achieved when it is proportional to the radius of the circles defining $\phi_{0}$. 

Since our prior uses fixed dictionaries, the approximation errors $e_{1}$ and $e_{2}$ are constant and calculated during the initialization phase. However, the correlation matrices $C_{1,t}$ and $C_{2,t}$ are dependent on the segmentation since they describe the general foreground and background distributions. Therefore, they need to be updated during the segmentation process. This is the equivalent of updating the foreground and background average intensity in the original Chan-Vese functional.

A time-step $t$ involves the following:
\begin{enumerate}
    \item Setting the narrow band $\left|\phi_{t}(x,y)\right|<\tau$ around the current boundary curve.
    \item Calculating $\kappa_{t}(x,y)$.
    \item Updating the correlation matrices $C_{1,t}$ and $C_{2,t}$ and the fidelity terms $\left(e_{1}^{T}C_{1,t}^{-1}e_{1}\right)$ and $\left(e_{2}^{T}C_{2,t}^{-1}e_{2}\right)$ according to $\phi_{t}(x,y)$. 
    \item Adding all terms with the functional parameters to get the `force' of the current step:
    \begin{equation}
    f_{t}(x,y)=\left[\mu\cdot\kappa_{t}(x,y)-\nu-\left(e_{1}^{T}C_{1,t}^{-1}e_{1}\right)+\left(e_{2}^{T}C_{2,t}^{-1}e_{2}\right)\right]\ .
    \end{equation}
    \item Calculating $dt$ to maintain the Courant-Friedrichs-Lewy (CFL) condition for numerical stability
        \begin{equation}
    dt=\frac{0.45}{\max\left(f_{t}(x,y)\right)}\ .
    \end{equation}
    \item Advancing $\phi$ by the force multiplied by $dt$
    \begin{equation} \phi_{t+1} = \phi_{t} + f_{t}dt\ . \end{equation}
    \item Performing re-initialization by the Sussman method \cite{sussman1994level} to ensure that $\phi(x,y)$ in the narrow band is a smooth SDF.
    \item Checking for convergence as described in Section \ref{sec:convergence}.
\end{enumerate}

\subsection{\label{sec:convergence} Convergence Check}
Let $S_t$ be the area of the foreground segments at time-step $t$:
\begin{equation} 
S_{t} = \underset{\Omega}{\int}H\left(\phi_{t}(x,y)\right)dxdy 
\end{equation}
Define $\Delta_{t}$ as:
\begin{equation} 
\Delta_{t} = S_{t} - S_{t-1} 
\end{equation}
We set a counter by checking if
\begin{equation} 
\frac{\Delta_{t}}{S_{t}} < d
\end{equation}
where $d$ is a parameter. In our experiments we set $d=0.005$. After 5 consecutive counts, convergence has been reached and we can stop the segmentation process.

\subsection{\label{sec:validation-classifier} Validation Classifier}
Let $P$ be a test patch and let $D_{1},D_{2}$ be 2 trained dictionaries of classes $G_{1 ,},G_{2}$, respectively.
Then
\begin{equation} \alpha_{1}=\text{OMP}\left(P,D_{1}\right)\,,\,\alpha_{2}=\text{OMP}\left(P,D_{2}\right) \end{equation}
such that
\begin{equation} P\approx D_{1}\alpha_{1}\,,\,P\approx D_{2}\alpha_{2} \end{equation}
We classify $P\in G_{1}$ if \begin{equation}\left\Vert P-D_{1}\alpha_{1}\right\Vert_{\Sigma_{1}}<\left\Vert P-D_{2}\alpha_{2}\right\Vert_{\Sigma_{2}} \end{equation} and $P\in G_{2}$ otherwise.

\subsection{Dictionary Training and Validation}
Training requires tuning training parameters, such as dictionary size, sparsity coefficient, number of iterations, as well as choosing initialization method, replacement of similar atoms and pursuit algorithm to use. For this research, KSVD was used, although the framework can be used with any other sparse dictionary learning methods, such as the Method of Optimal Directions (MOD) \cite{engan1999method}, Online Dictionary Learning (ODL) \cite{mairal2009online}, Fisher Discrimination Dictionary Learning (FDDL) \cite{yang2011fisher}, and `deep dictionary learning' \cite{tariyal2016deep}. The output of this step are two dictionaries, one for the object and one for the background. \par

\begin{figure}[H]
     \centering
     \begin{subfigure}[t]{0.45\textwidth}
         \centering
         \includegraphics[width=\textwidth]{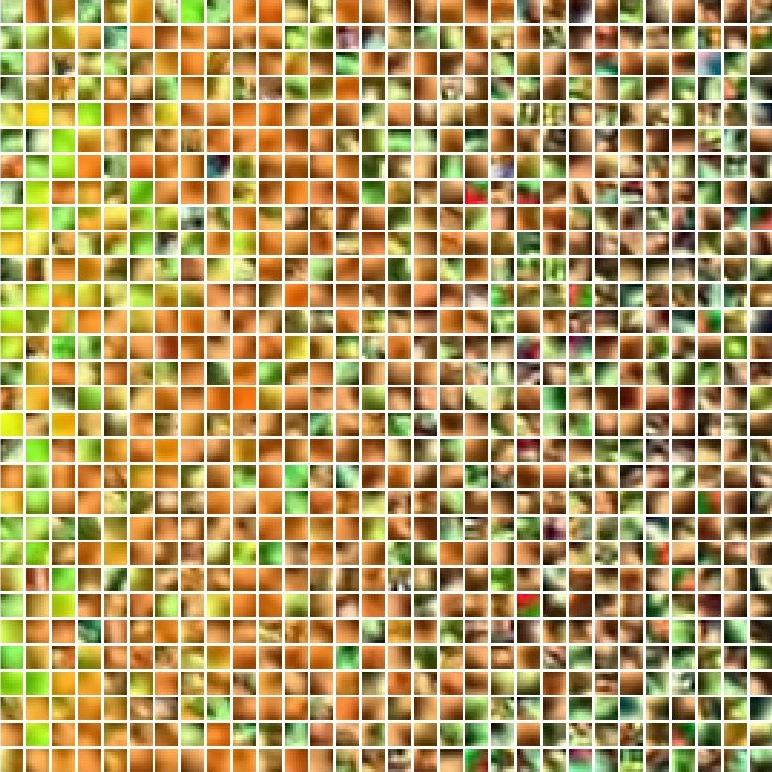}
         \caption{Dictionary of Infected plants}
         \label{fig:infected}
     \end{subfigure}
     \hfill
     \begin{subfigure}[t]{0.45\textwidth}
         \centering
         \includegraphics[width=\textwidth]{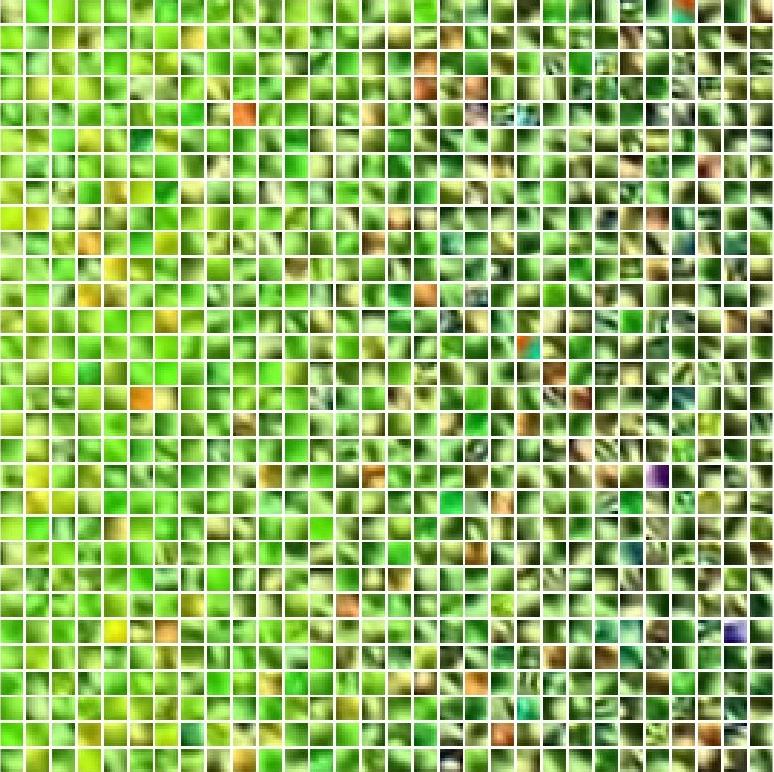}
         \caption{Dictionary of Healthy plants}
         \label{fig:healthy}
     \end{subfigure}
    \caption{Dictionaries trained with KSVD according to the process in \ref{sec:level-set-ksvd-framework}.}
        \label{fig:Brown-and-green}
\end{figure}

The trained dictionaries are shown in Figure \ref{fig:Brown-and-green}. Both dictionaries are sorted column-wise by brightness. Top-left atom is the brightest, going downward. Bottom-right atom is the darkest.
\par
Dictionary for the Infected plants presented in Figure \ref{fig:infected}. The atoms of the Infected dictionary predominantly contains patches of dead zones and barren land, depicted as brown segments. However, it contains some healthy patches, as shown by the green segments. This is expected, since some infected areas have a few healthy plants, particularly around the edges.
\par
Dictionary for the Healthy plants presented in Figure \ref{fig:healthy}. It contains mostly healthy patches with strong green hues. Note that the atoms have a wide range of hues and patterns to capture different variety and growth stages of plants.
\par

To validate the dictionaries we created a simple classifier, presented in Section \ref{sec:validation-classifier}. The classifier mimics the level-set functional by approximating a test patch with both dictionaries and classifying according to a minimal $L_2$-norm. \par

\begin{figure}[H]
\noindent \begin{centering}
\includegraphics[scale=0.25]{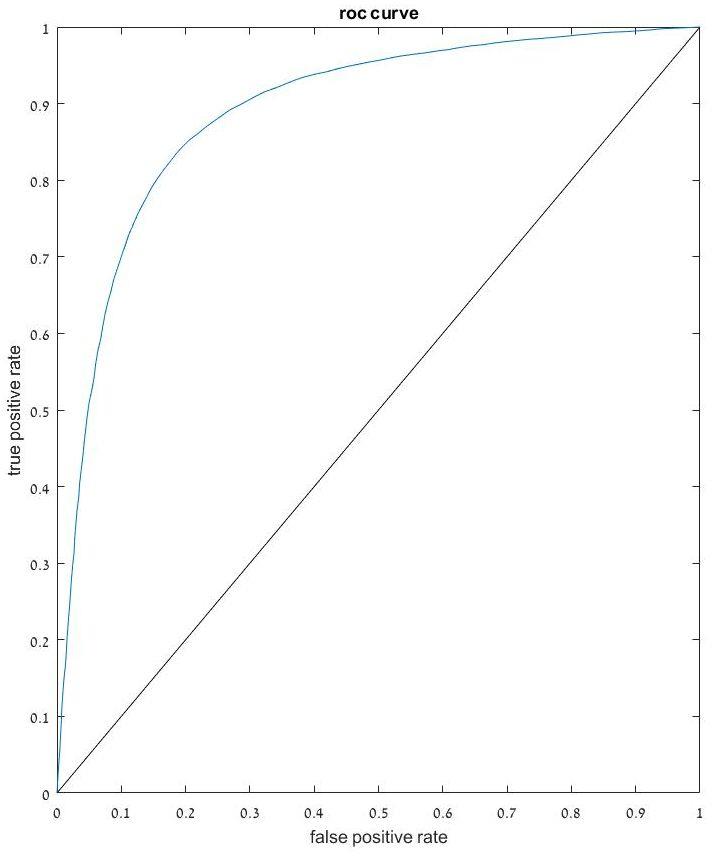}
\par\end{centering}
\caption{\label{fig:ROC-curve-and}ROC curve to validate the classifier created with infected and healthy dictionaries as described in Section \ref{sec:validation-classifier}.}
\end{figure}

During the refinement process of the dictionaries, a ROC curve (see \ref{sec:ROC-Curve}) was generated to analyse the quality and the validity of the models used. It is presented in Figure \ref{fig:ROC-curve-and}. The curve shows that a high quality model has been achieved, as depicted by the closeness of the curve to the (0,1) point. An excellent Area-Under-the-Curve (AUC) of greater than 0.9 suggests that the level of false positives is minimal.

\subsection{Data acquisition}
The data for this research was acquired in a method mimicking a farmer scouting his field: First, an anomaly was detected by manual scouting. Then, a UAV mounted with a digital camera was programmed to take images on a flight course covering the entire field. Afterwards, the images were downloaded to a memory card and transferred for research and computations. The camera mounted was a Mirror-less Sony alpha6000.

The datasets generated and analysed during the current study are available from the corresponding author on reasonable request.

\subsection{System used}
All simulations and computations were conducted on a personal computer Intel i7-4790 @ 3.60 GHZ , 16GB RAM with Windows 7 SP1 64-bit on Matlab R2016b. 
Supplementary simulations were conducted on a virtual server with similar capabilities, with Windows Server 2012 and Matlab 2016b.
For accelerating KSVD computation, code from Ron Rubinstein and Michael Elad \cite{Rubinstein2008} was used.

\section{\label{sec:level-set-ksvd-framework} Macrophomina segmentation}
To demonstrate our method on real-life data, we collected a set of agricultural aerial imagery of cotton field with fungi infection. Our demonstration includes a semi-automatic framework, which requires creating dictionaries for a single image, and using it to segment the other images of the same set. In our set, all images are similar since they were taken at the same place, with similar conditions.

In general, given a set of similar images, we can use one of them to segment all the others. This method can replace pre-trained generalist segmentation algorithms. In addition, preparation is short and simple since the ``training set'' is just a single image.
Our semi-automatic framework for segmentation involves the following steps: 
\\

\fbox{\parbox{\textwidth}{
\begin{enumerate}
\item Create 2 datasets of patches for foreground and background
\item Train and validate dictionaries for each dataset, using a sparse dictionary learning method (e.g. KSVD)
\item Segment the other images by minimizing the level-set KSVD functional
\end{enumerate}}}
%
\subsection{Dataset Creation} 
The steps for dataset creation are: 
\\

\fbox{\parbox{\textwidth}{
\begin{enumerate}[noitemsep,topsep=0pt]
\item Choose an image with distinct infected and healthy areas.
\item Mark large areas as infected or healthy to create 2 masks.
\item Extract patches from each mask. Balance if needed.
\item Shuffle infected and healthy patches together and split with 70-30 ratio for training set and test set, respectively.
\end{enumerate}}}\\
\\

An important training parameter is the patch size, which also determines the atom sizes. Tuning this parameter requires creating new datasets and repeating the entire process. In our experiments, we tried several options. The best results were achieved with 8-by-8 RGB patches. \par

Another important caveat is balancing between the classes. In our experiments, the images have dominant healthy areas and few infected. We used random under-sampling of healthy patches to force 1-to-1 ratio between the classes in the training dataset.


\subsection{Segmentation}
The results obtained through carrying out the described process are summarised here alongside a critical analysis.

Segmentation can be implemented as follows: 
\\

\fbox{\parbox{\textwidth}{
\begin{enumerate} [topsep=0pt]
\item Create patch per pixel (Set $M=\left\{ P_{n}\left(x,y\right)\right\} _{n=1}^{N}$).
\item Compute the sparse representation $\alpha_{1}(x,y)$ and $\alpha_{2}(x,y)$ for all patches with Batch-OMP (See \cite{Rubinstein2008}).
\item Compute error approximations  $e_{1}(x,y),e_{2}(x,y)$ for all patches (see Appendix \ref{appendix:fast-compute} for details)
\item Initialize $\mu$, $\nu$ and $\phi_{0}$
\item Minimize functional \eqref{eq:main-cvd} until convergence
\end{enumerate}}}
\\

This implementation is fast due to computing the feature vectors $e_{1},e_{2}$ for all pixels before minimizing the functional. During the minimization process, we only update the correlation matrices. \par
We used the following Chan-Vese parameters: $\mu=25\,,\,\nu=35$

Post-processing concludes the segmentation. Post-processing can include, for example, removing small segmented areas, smoothing the curves and merging nearby segmented areas (e.g. with morphology  \cite{trove.nla.gov.au/work/13090689}).
In our experiments we removed a few false segments that were created around bright-colored flowers.


\subsection{Segmentation Results}

\begin{figure}[H]
     \centering
     \begin{subfigure}[t]{0.45\textwidth}
         \centering
         \includegraphics[width=\textwidth]{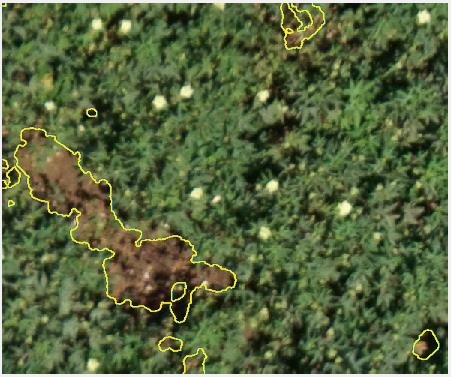}
         \caption{Image with one large segment and few small ones}
         \label{fig:one-large}
     \end{subfigure}
     \hfill
     \begin{subfigure}[t]{0.45\textwidth}
         \centering
         \includegraphics[width=\textwidth]{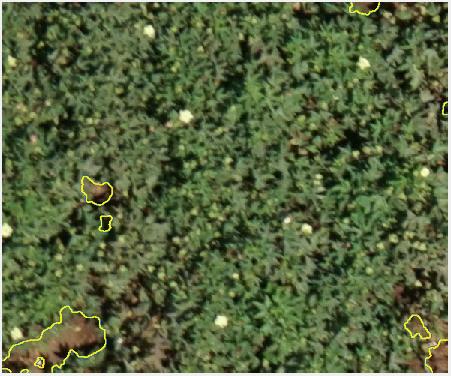}
         \caption{Image with a medium segment and few small segments}
         \label{fig:one-medium}
     \end{subfigure}
     \hfill
     \begin{subfigure}[t]{0.45\textwidth}
         \centering
         \includegraphics[width=\textwidth]{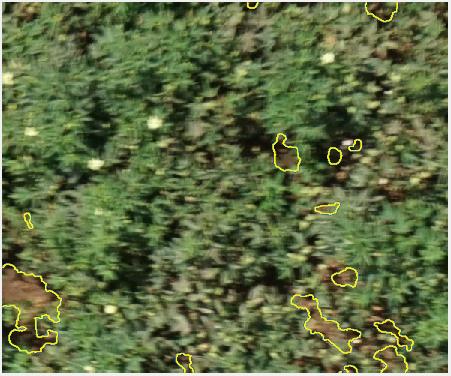}
         \caption{Blurred image with many segments}
         \label{fig:blurred}
     \end{subfigure}
     \hfill
     \begin{subfigure}[t]{0.45\textwidth}
         \centering
         \includegraphics[width=\textwidth]{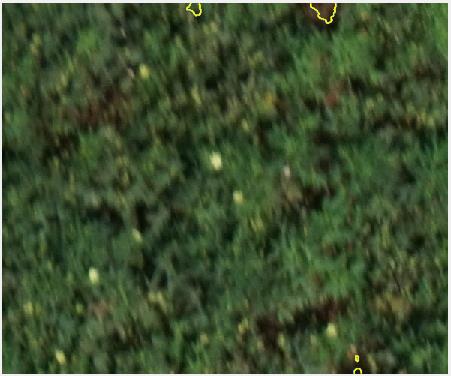}
         \caption{Very blurred image with tiny segments}
         \label{fig:very-blurred}
     \end{subfigure}
    \caption{Segmentation results for \textit{Macrophomina} using the proposed method. Infected areas are surrounded by yellow contour}
        \label{fig:Sample-segmentation-results}
\end{figure}

We segment multiple acquired images using Level-set KSVD algorithm in order to assess its ability to outline the healthy and infected areas of the image. Example outputs of this are shown in Figure \ref{fig:Sample-segmentation-results}, where the infected areas are surrounded by a yellow contour. The algorithm was used on a range of different crop field images, with varying levels of infected areas and image quality. As can be seen in the figure, Level-set KSVD is able to successfully segment high variability of infected areas.

Figure \ref{fig:one-large} shows the algorithm's ability to identify a large segment of infected area alongside multiple smaller ones. Here, the large infected area has a particularly irregular shape, where the contour is successfully identified without erroneous spikes. \par

Figure \ref{fig:one-medium} is an example of an area with one medium and multiple smaller infected segments, and this too, is able to be successfully classified by the algorithm. The boundaries of the infected areas are identified accurately with no spikes nor deviations. It can also be seen that even relatively smaller infected areas are successfully captured. \par

In Figure \ref{fig:blurred}, which is an example of a blurred image with many scattered infected segments, it can be seen that, despite the optical challenge, the algorithm is able to accurately identify the contours of the infected areas. Here, a range of area sizes are present, and this does not interfere with the level of accuracy. This is a particularly powerful example, as it illustrates the algorithm's robustness in real world applications where images may not be perfectly focused, or may be subjected to motion blur which originates from the UAV dynamics. \par

Lastly, Figure \ref{fig:very-blurred} highlights the limitations of the algorithm, where an extremely blurred image input is no longer able to be segmented without mistakes. Here, instead of incorrectly classifying non infected areas as infected, the algorithm does not classify the infected areas at all (which are difficult to identify even by experts). This indicates that the false positive rate is low. \par

In summary, the proposed Level-set KSVD can successfully detect infected areas of different sizes and has no falsely segmented areas (as shown by the four examples in Figure \ref{fig:Sample-segmentation-results}). It identifies the contours of the areas well, capturing the boundary accurately which is paramount for the use case where only the infected areas should be treated.
The algorithm produces excellent segmentation results for sharp images (as shown in Figures \ref{fig:one-large} and \ref{fig:one-medium}), and has satisfactory results in blurred images, which are traditionally challenging to analyse (Figure \ref{fig:blurred}). A very blurred image, for example Figure \ref{fig:very-blurred}, is not able to be as successfully classified, however, it shows no erroneous false positives, which is critical for the use case.
Overall, the Level-set KSVD generalizes well for different field situations and image qualities. This is shown by having good results for all examples shown in Figure \ref{fig:Sample-segmentation-results}, while using the same dictionaries and Chan-Vese parameters.
All of these observations demonstrate the power of Level-set KSVD and its applicability to this use case. \par

\subsection{Comparison to other methods}
\begin{figure}[H]
     \centering
     \begin{subfigure}[t]{0.45\textwidth}
         \centering
         \includegraphics[width=\textwidth]{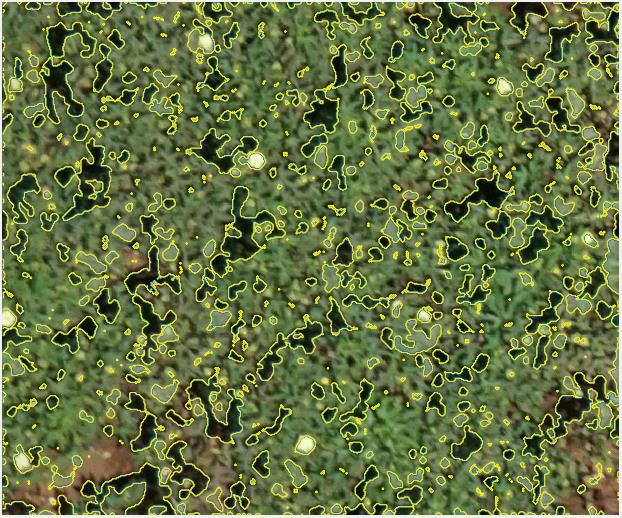}
         \caption{Vector-valued Chan-Vese \cite{Chan2000}}
         \label{fig:Vector-valued}
     \end{subfigure}
     \hfill
     \begin{subfigure}[t]{0.45\textwidth}
         \centering
         \includegraphics[width=\textwidth]{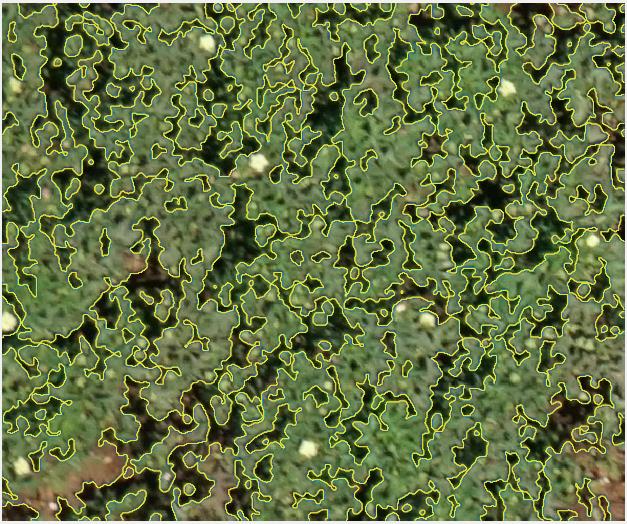}
         \caption{Vector-valued Chan-Vese with Gabor filters \cite{sandberg2002level}}
         \label{fig:Vector-valued Gabor}
     \end{subfigure}
     \hfill
     \begin{subfigure}[t]{0.45\textwidth}
         \centering
         \includegraphics[width=\textwidth]{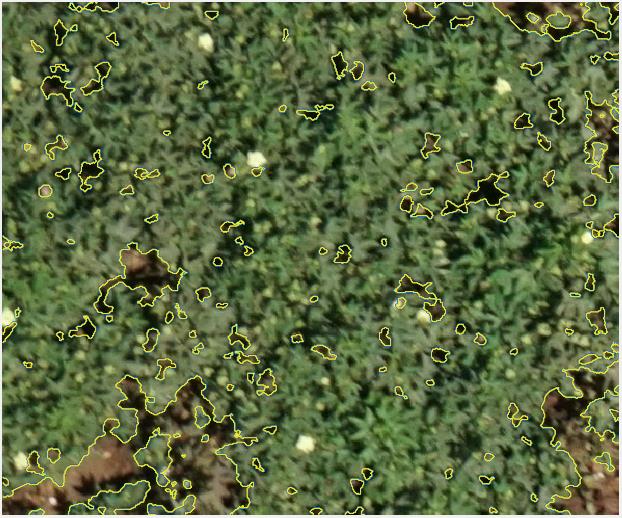}
         \caption{Basic Chan-Vese with a{*}-channel from CIELAB color space \cite{CIELAB1976}}
         \label{fig:a-channel}
     \end{subfigure}
     \hfill
     \begin{subfigure}[t]{0.45\textwidth}
         \centering
         \includegraphics[width=\textwidth]{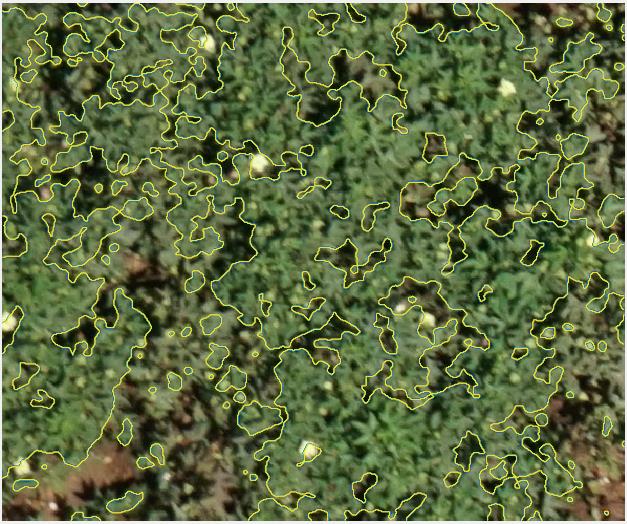}
         \caption{Same as \ref{fig:a-channel}, with Gabor filters as in \ref{fig:Vector-valued Gabor}}
         \label{fig:a-channel Gabor}
     \end{subfigure}
     \hfill
     \begin{subfigure}[t]{0.45\textwidth}
         \centering
         \includegraphics[width=\textwidth]{9609_400-900_200-800_ksvd.jpg}
         \caption{Our proposed Level-set KSVD}
         \label{fig:proposed}
     \end{subfigure}
     \hfill
     \begin{subfigure}[t]{0.45\textwidth}
         \centering
         \includegraphics[width=\textwidth]{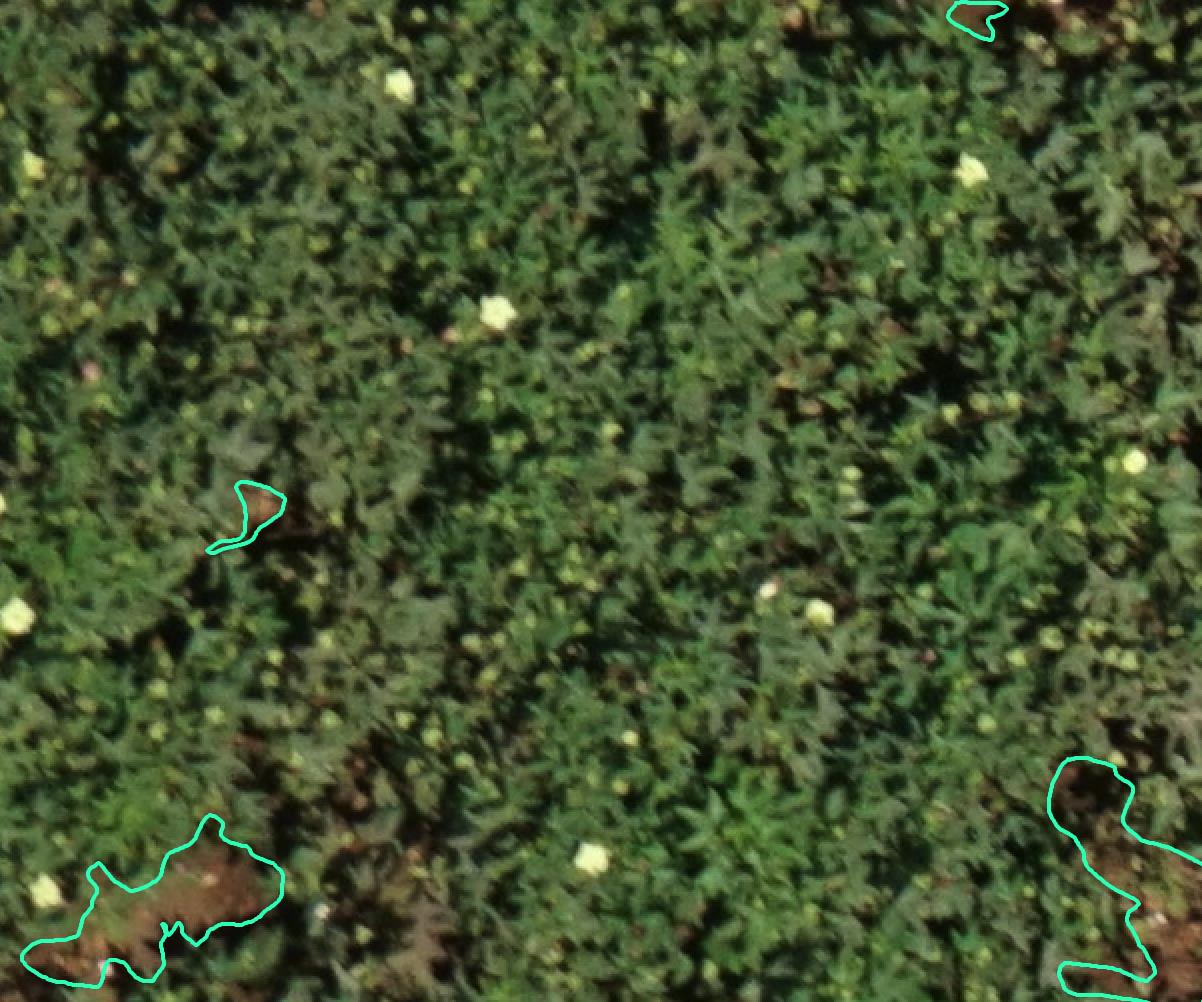}
         \caption{Human annotated (ground truth) segmentation}
         \label{fig:GT}
     \end{subfigure}
        \caption{Comparison between different segmentation methods. Infected areas are surrounded by yellow contour}
        \label{fig:regular-chanvese-comparison}
\end{figure}

In order to further evaluate the Level-Set KSVD method, results were compared to those obtained using four different approaches. These are presented in Figure \ref{fig:regular-chanvese-comparison} alongside the Level-Set KSVD method as well as a manually annotated ground truth for reference.\par

Figure \ref{fig:Vector-valued} shows a baseline segmentation result. Basic Chan-Vese scheme for vector-valued images is used (as proposed in \cite{Chan2000}). This approach predominantly segments the dark areas in the image and has a lot of noise. \par
In order to deal with the noise, a Chan-Vese scheme with Gabor filters was used, as suggested in \cite{sandberg2002level} and this is shown in Figure \ref{fig:Vector-valued Gabor}. This method (like the previous) segments based on brightness, however it has less noise and displays better clustering of neighbouring segments. \par
In order to achieve improved segmentation between healthy and infected areas, a hand-crafted simple solution of changing the colour space was utilized. Here, the CIELAB colour space was used (as defined in \cite{CIELAB1976}), which is designed to approximate the human vision using three channels: L*, a* and b* for lightness, red-green spectrum and blue-yellow spectrum, respectively. The red-green spectrum (a* channel), provides excellent separation between brown and green hues. As such, the a* channel was isolated as a gray-scale image and was segmented using a basic Chan-Vese schema. This approach provides improved results, successfully segmenting the healthy and infected areas. It does however, suffer from over segmentation and noisy regions as can be seen in Figure \ref{fig:a-channel}. \par
Finally, to overcome the noise, the a* channel approach was combined with Gabor filters, as shown in Figure \ref{fig:a-channel Gabor}. The result of this was less favourable, where the segmentation again was based on brightness. \par
Figure \ref{fig:proposed} shows a successful segmentation using the proposed Level-Set KSVD. \par
Figure \ref{fig:GT} is manually annotated ground truth for reference. When comparing this to the segmentation achieved using the Level-Set KSVD method, it can be seen that the areas identified are extremely similar, again, alluding to its applicability and value in this use case.  \par

In conclusion, it can be seen out the Level-Set KSVD outperforms the other methods when comparing to the ground-truth annotation. The Level-Set KSVD is able to segment healthy and infected areas and does not generate noise. \par

\section{Conclusions}

In this work, a novel model that we name \textbf{level-set KSVD} is presented. The model combines the advantages of dictionary learning and variational segmentation. It needs only a single manually annotated image in contrast to some learning methods that require many labeled/segmented images to perform well. 
Using dictionaries enables to include the benefit of learning `neighborhood' features and delivers segmentation results superior to other methods. The model was derived mathematically and tested on real-life agricultural images to detect spread of  \textit{Macrophomina}.
Future work includes: applying the model to `difficult' images (with many features and elements), improving the model by incorporating models such as multi-phase Chan-Vese \cite{Vese2002} , color KSVD \cite{Mairal2008} or multi-scale KSVD \cite{Mairal2008b}, and generalizing the model to other variational methods and testing its results with more dictionary learning methods.


\bibliographystyle{IEEEtran}
\bibliography{Level_set_ksvd.bib}

\begin{appendices}
\section{Patch-features equivalence}
\label{appendix:simplify}
Instead of viewing the patch centered at a pixel as `neighborhood' features of this pixel, we can think of each patch as a small image and separate it to object part and background part. Then we compare each part to its approximation to get the segmentation penalty. A functional for this view can be defined as follows:
\noindent 
\begin{align}
F_{\mu,\nu,\lambda_{1},\lambda_{2}}\left(\phi,\alpha_{1},\alpha_{2}\right) = 
 & \underset{\Omega\left(x,y\right)}{\int}\mu\delta\left(\phi\left(x,y\right)\right)\cdot\left|\nabla\left(\phi\left(x,y\right)\right)\right|
 +\nu H\left(\phi\left(x,y\right)\right)\nonumber \\
 & +\lambda_{1}\left|P\left(x,y\right)\cdot H\left(\phi\left(x,y\right)\right)-\sum_{k=1}^{K_{1}}\alpha_{1k}D_{1k}H\left(\phi\left(x,y\right)\right)\right|^{2}\nonumber\label{eq:initial-functional} \\
 & +\lambda_{2}\left|P\left(x,y\right)\cdot\left(1-H\left(\phi\left(x,y\right)\right)\right)-\sum_{k=1}^{K_{2}}\alpha_{2k}D_{2k}\left(1-H\left(\phi\left(x,y\right)\right)\right)\right|^{2} \nonumber \\
 & dxdy
\end{align}

where $P\left(x,y\right)$ is a patch centered around $\left(x,y\right)$ coordinate and each patch is approximated by a dictionary $D_i$ and coefficient vector $\alpha_i$, where $i=1,2$ indicates the object and background dictionaries. \\ 
Notice the Heaviside function is applied to patches and approximations separately. Therefore, as expected, only pixels in the patch that belong to the object are compared with the object approximation (and the same goes for background pixels in the patch).

Now we show that this functional is equivalent to Equation \eqref{eq:first-functional}. Recall that Equation \eqref{eq:first-functional} defines a functional that indicates each patch $P\left(x,y\right)$ is segmented within itself into object and background using the dictionaries. \\

Since $H\left(z\right)$ is a binary function, we have:
\begin{equation}
\left|H\left(z\right)\right|^{2}=\left|H\left(z\right)\right|=H\left(z\right)
\end{equation}
Similarly, for the background term we have:
\begin{equation}
\left|1-H\left(z\right)\right|^{2}=\left|1-H\left(z\right)\right|=1-H\left(z\right)
\end{equation}
\noindent
By factoring out, and using the above relations, we get:
\begin{align*}
\left|P\left(x,y\right)\cdot H\left(\phi\left(x,y\right)\right)-\sum_{k=1}^{K_{1}}\alpha_{1k}D_{1k}H\left(\phi\left(x,y\right)\right)\right|^{2} & =\\
\left(\left|P\left(x,y\right)-\sum_{k=1}^{K_{1}}\alpha_{1k}D_{1k}\right|\cdot\left|H\left(\phi\left(x,y\right)\right)\right|\right)^{2} & =\\
\left|P\left(x,y\right)-\sum_{k=1}^{K_{1}}\alpha_{1k}D_{1k}\right|^{2}\cdot\left|H\left(\phi\left(x,y\right)\right)\right|^{2} & =\\
\left|P\left(x,y\right)-\sum_{k=1}^{K_{1}}\alpha_{1k}D_{1k}\right|^{2}H\left(\phi\left(x,y\right)\right)
\end{align*}
The dictionary approximation $\sum_{k=1}^{K_{1}}\alpha_{1k}D_{1k}$
can be viewed as a matrix multiplication of the dictionary matrix
$D_{1}$ with the corresponding row $\alpha_{1}\left(x,y\right)$
in the sparse representation matrix $A_{1}$:
\begin{equation}
\sum_{k=1}^{K_{1}}\alpha_{1k}D_{1k}=D_{1}\alpha_{1}\left(x,y\right)
\end{equation}
Putting this in our expression, and omitting $\left(x,y\right)$ for readability, we get:
\begin{equation}
\left|P-\sum_{k=1}^{K_{1}}\alpha_{1k}D_{1k}\right|^{2}H\left(\phi\right)=\left|P-D_{1}\alpha_{1}\right|^{2}H\left(\phi\right)
\end{equation}
Following the same arguments and omissions for $D_{2}$, we get:
\begin{equation}
\begin{aligned}
\left|P\cdot\left(1-H\left(\phi\right)\right)-\sum_{k=1}^{K_{2}}\alpha_{2k}D_{2k}\left(1-H\left(\phi\right)\right)\right|^{2} = \\
\left|P-D_{2}\alpha_{2}\right|^{2}\left(1-H\left(\phi\right)\right)
\end{aligned}
\end{equation}
Assigning the terms derived above in Equation \eqref{eq:initial-functional} gives exactly Equation \eqref{eq:first-functional}.

\section{Diagonal of matrix product}
\label{appendix:fast-compute}
We would like to calculate error approximations for all pixels in the image for both dictionaries. 
These are the terms $\left\|P(x,y)-D_{i}\alpha_{i}(x,y)\right\|^{2}_{\Sigma_{i}}, i=\{1,2\}$ for all $\left(x,y\right)\in\Omega$, where $\Omega$ is the image region on the image plane. \par
Since this can be computationally expensive, we can considerably reduce the computation time using some mathematical relations. \par

First, recall our definition of approximation error (see Equation \eqref{eq:approx-error}):
\begin{equation} 
\label{eq:approx-error-recall}
e_{1}\left(x,y\right)=P\left(x,y\right)-D_{1}\alpha_{1}\left(x,y\right) \end{equation}
Therefore, we need to calculate
\begin{equation} e_{1}^{T}\Sigma_{1}^{-1}e_{1}  ,  e_{2}^{T}\Sigma_{2}^{-1}e_{2} \end{equation}
 for all $\left(x,y\right)\in\Omega$. \par
By listing all  $\left(x,y\right)$ approximation errors from Equation \eqref{eq:approx-error-recall}, we get a system of linear equations. We can re-write this system as a matrix equation:
\begin{equation} E_{1}=P-D_{1}A_{1} \end{equation}
Using the matrix notation, and our statistical approach (see Equation \eqref{eq:stat-approach}), we get the following:
\begin{equation} \left\|P-D_{1}A_{1}\right\|^{2}_{\Sigma_{1}}=E_{1}^{T}\Sigma_{1}^{-1}E_{1} \end{equation}
Note that $E_{1}=\{e_1(x,y) | (x,y)\in \Omega\}$ is considered as an ordered set e.g a large matrix such that: 
\begin{equation} E_{1}\in\mathbb{R}^{\left|\Omega\right|\times k}. \end{equation}
For our purposes, we only need the diagonal of this matrix:
\begin{equation} diag\left(E_{1}^{T}\Sigma_{1}^{-1}E_{1}\right) \end{equation}
Note that the diagonal of a matrix product can be written with element-wise products (Hadamard products) and summation-over-rows:
\begin{equation} diag\left(A B \right) = \Sigma_{j=1}^{n}\left(A \odot B^{T} \right)_{ij} \end{equation}
Finally we arrive at the fast-computation diagonal:
\begin{equation} diag\left(E_{1}^{T}\Sigma_{1}^{-1}E_{1}\right) = \Sigma_{j=1}^{n}\left( \left(E_{1}^{T}\Sigma_{1}^{-1}\right)\odot E_{1}^{T} \right)_{ij}  \end{equation}

\section{Chan-Vese log-likelihood}
\label{appendix:chan-vese-log-likelihood}

Assume single-variable Gaussian distribution with random variable $x$ mean $\mu$ and variance $\sigma$:

\begin{equation}
f\left(x,\mu,\sigma\right)=\frac{1}{\sqrt{2\pi\sigma^{2}}}\exp\left(-\frac{1}{2}\left(\frac{x-\mu}{\sigma}\right)^{2}\right)
\end{equation}

Now, take log to get the fidelity term to minimize:

\begin{equation}
\ln\left(f\right)=\ln\left(\frac{1}{\sqrt{2\pi\sigma^{2}}}\right)-\frac{1}{2\sigma^{2}}\left(x-\mu\right)^{2}
\end{equation}

The first term is a constant which does not change the minimum argument.
in the second term, if we define $\lambda=\frac{1}{2\sigma^{2}}$ and
we change $x$ to be the pixel intensity value $u_{0}$ and the mean
to be the average intensity $c$ , we get the fidelity terms for ``inside
contour'' and ``outside contour'' in Chan-Vese functional:

\begin{equation}
-\lambda_{1}\left(u_{0}-c_{1}\right)^{2}+\lambda_{2}\left(u_{0}-c_{2}\right)^{2}
\end{equation}

We can generalize this to multi-variate Gaussian distribution with random vector $x\in\mathbb{R}^{n}$ , vector mean $\mu$ and covariance matrix $\Sigma$:

\begin{equation}
f\left(x,\mu,\Sigma\right)=\left(2\pi\right)^{-\frac{k}{2}}\det\left(\Sigma\right)^{-\frac{1}{2}}\exp\left(-\frac{1}{2}\left(x-\mu\right)^{T}\Sigma^{-1}\left(x-\mu\right)\right)
\end{equation}

After taking log:

\begin{equation}
\ln\left(f\right)=-\frac{k}{2}\ln\left(2\pi\right)-\frac{1}{2}\ln\left(\det\left(\Sigma\right)\right)-\frac{1}{2}\left(x-\mu\right)^{T}\Sigma^{-1}\left(x-\mu\right)
\end{equation}

As with the univariate case, the first terms are constants, and the last term gives the fidelity term of the vector-valued Chan-Vese variant.

\section{Correlation matrix}
\label{appendix:correlation-matrix}

We will now derive the fidelity term assuming Gaussian distribution with unit variance.
Let $\mathbf{X}$ be some random vector variable with entries $X_{1}, X_{2}, ..., X_{n}$.
Let $\Sigma(\mathbf{X})$ be the covariance matrix of $\mathbf{X}$.
Define $C(\mathbf{X})$ as the correlation matrix of $\mathbf{X}$ by the following equation:
\begin{equation}
C(\mathbf{X})=diag\left(\Sigma(\mathbf{X})\right)^{-\frac{1}{2}}\cdot\Sigma(\mathbf{X})\cdot diag\left(\Sigma(\mathbf{X})\right)^{-\frac{1}{2}}
\end{equation}

where $diag\left(\Sigma(\mathbf{X})\right)$ is a diagonal matrix of the variances $\sigma(X_{i}), i=1,...,n$.

Equivalently, the correlation matrix is the covariance matrix of the standardized random variable. Each variable $X_{i}$ in $\mathbf{X}$ is scaled to variance of $1$ and the covariances are scaled accordingly:

\begin{equation}\operatorname{C}(\mathbf{X})=\begin{bmatrix}1 & {\frac{\operatorname{E}[(X_{1}-\mu_{1})(X_{2}-\mu_{2})]}{\sigma(X_{1})\sigma(X_{2})}} & \cdots & {\frac{\operatorname{E}[(X_{1}-\mu_{1})(X_{n}-\mu_{n})]}{\sigma(X_{1})\sigma(X_{n})}}\\
\\
{\frac{\operatorname{E}[(X_{2}-\mu_{2})(X_{1}-\mu_{1})]}{\sigma(X_{2})\sigma(X_{1})}} & 1 & \cdots & {\frac{\operatorname{E}[(X_{2}-\mu_{2})(X_{n}-\mu_{n})]}{\sigma(X_{2})\sigma(X_{n})}}\\
\\
\vdots & \vdots & \ddots & \vdots\\
\\
{\frac{\operatorname{E}[(X_{n}-\mu_{n})(X_{1}-\mu_{1})]}{\sigma(X_{n})\sigma(X_{1})}} & {\frac{\operatorname{E}[(X_{n}-\mu_{n})(X_{2}-\mu_{2})]}{\sigma(X_{n})\sigma(X_{2})}} & \cdots & 1
\end{bmatrix}\end{equation}
Recall the Gaussian distribution function:
\begin{equation}
f\left(X,\mu,\Sigma\right)=\frac {e^{-{\frac {1}{2}}({\mathbf {X} }-{\boldsymbol {\mu }})^{\mathrm {T} }{\boldsymbol {\Sigma }}^{-1}({\mathbf {X} }-{\boldsymbol {\mu }})}}{\sqrt {(2\pi )^{k}|{\boldsymbol {\Sigma }}|}}
\end{equation}

written in a different way:

\begin{equation}
f\left(X,\mu,\Sigma\right)=\left(2\pi\right)^{-\frac{k}{2}}\det\left(\Sigma\right)^{-\frac{1}{2}}\exp\left(-\frac{1}{2}\left(X-\mu\right)^{T}{\Sigma}^{-1}\left(X-\mu\right)\right)
\end{equation}

Assuming random variables with unit-variance and zero mean simply removes $\mu$ and replaces $\Sigma$ with $C$:
\begin{equation}
f\left(X,\mu=0,\Sigma=C\right)=\left(2\pi\right)^{-\frac{k}{2}}\det\left(C\right)^{-\frac{1}{2}}\exp\left(-\frac{1}{2}\left(X-\mu\right)^{T}C^{-1}\left(X-\mu\right)\right)
\end{equation}

After taking log:
\begin{equation}
\ln\left(f\right)=-\frac{k}{2}\ln\left(2\pi\right)-\frac{1}{2}\ln\left(\det\left(C\right)\right)-\frac{1}{2}\left(X-\mu\right)^{T}C^{-1}\left(X-\mu\right)
\end{equation}

This is closely related to the regular form since $C$ is simply a scaled $\Sigma$:
\begin{align}
\ln\left(f\right) = 
& -\frac{k}{2}\ln\left(2\pi\right) )\nonumber\\
& -\frac{1}{2}\ln\left(\det\left(diag\left(\Sigma\right)^{-\frac{1}{2}}\cdot\Sigma\cdot diag\left(\Sigma\right)^{-\frac{1}{2}}\right)\right) )\nonumber\\
& -\frac{1}{2}\left(X-\mu\right)^{T}\left(diag\left(\Sigma\right)^{-\frac{1}{2}}\cdot\Sigma\cdot diag\left(\Sigma\right)^{-\frac{1}{2}}\right)^{-1}\left(X-\mu\right) )\nonumber\\
\end{align}

Denoting the variance of element $x_{i}$ with $\sigma_{i}$, the diagonal matrix $diag\left(\Sigma\right)$ is:

\begin{equation}diag\left(\Sigma\right)(X)=\begin{bmatrix}\sigma_{1} & 0 & \cdots & 0\\
\\
0 & \sigma_{2} & \cdots & 0\\
\\
\vdots & \vdots & \ddots & \vdots\\
\\
0 & \cdots & 0 & \sigma_{n}
\end{bmatrix}\end{equation}

Therefore, the diagonal matrix $diag\left(\Sigma\right)^{\frac{1}{2}}$ is:

\begin{equation}diag\left(\Sigma\right)^{\frac{1}{2}}(X)=\begin{bmatrix}\frac{1}{\sqrt{\sigma_{1}}} & 0 & \cdots & 0\\
\\
0 & \frac{1}{\sqrt{\sigma_{2}}} & \cdots & 0\\
\\
\vdots & \vdots & \ddots & \vdots\\
\\
0 & \cdots & 0 & \frac{1}{\sqrt{\sigma_{n}}}
\end{bmatrix}\end{equation}

Denoting $\sigma=\left\{ \sigma_{1},\sigma_{2},\ldots,\sigma_{n}\right\}$ and putting it in the previous equation, we get the following Chan-Vese log-likelihood fidelity term:

\begin{equation}
\ln\left(f\right)=const-\frac{1}{2}\left(\frac{X-\mu}{\sqrt{\sigma}}\right)^{T}\cdot\Sigma^{-1}\cdot\left(\frac{X-\mu}{\sqrt{\sigma}}\right)
\end{equation}

Note is it similar to the fidelity term we received from assuming Gaussian distribution, except dividing by the standard deviation $\sqrt{\sigma}$ for $X-\mu$.

\end{appendices}

\end{document}